\documentclass[runningheads]{llncs}

\usepackage{eccv}
\usepackage{eccvabbrv}
\usepackage{graphicx}
\usepackage{booktabs}
\usepackage{tabularx}
\usepackage{array}
\usepackage{amsmath}
\usepackage{hyperref}
\usepackage[accsupp]{axessibility}

\newcolumntype{Y}{>{\raggedright\arraybackslash}X}
\graphicspath{{figures/}}
\begin{document}

\title{The 10th AI City Challenge}
\titlerunning{The 10th AI City Challenge}

\author{Zheng Tang\inst{1} \and Shuo Wang\inst{1} \and David C. Anastasiu\inst{2} \\ 
Ming-Ching Chang\inst{3} \and Anuj Sharma\inst{4} \and Quan Kong\inst{5} \\ 
Munkhjargal Gochoo\inst{6} \and Jun-Wei Hsieh\inst{7} \and Tomasz Kornuta\inst{1} \\ 
Zhedong Zheng\inst{8} \and Renran Tian\inst{9} \and Judah Goldfeder\inst{10} \\ 
Fulgencio Navarro\inst{11} \and Yuxing Wang\inst{1} \and Yizhou Wang\inst{1} \\ 
Sameer Satish Pusegaonkar\inst{1} \and Anqi Li\inst{1} \and Nalin Dadhich\inst{1} \\ 
Ridham Kachhadiya\inst{2} \and Dhanishtha Patil\inst{2} \and Haoquan Liang\inst{1} \\ 
Jiajun Li\inst{1} \and Han Zhang\inst{1} \and Yilin Zhao\inst{1} \\ 
Zaid Pervaiz Bhat\inst{1} \and Shuyu Yang\inst{12} \and Ashutosh Kumar\inst{5} \\ 
Rong Wang\inst{5} \and Rafael Martin Nieto\inst{11} \and Peter Christiansen\inst{11} \\ 
Ahmed Abduljawad\inst{6} \and Mohanrasu Shanmugam\inst{6} \and Nadeem Shaik\inst{6} \\ 
Sujit Biswas\inst{1} \and Xunlei Wu\inst{1} \and Vidya Murali\inst{1} \\ 
\and Rama Chellappa\inst{13}}
\authorrunning{Z. Tang et al.}
\institute{\footnotesize
\begin{tabular}{@{}c@{}}
\textsuperscript{1}\,NVIDIA \quad
\textsuperscript{2}\,Santa Clara University \\
\textsuperscript{3}\,University at Albany, SUNY \quad
\textsuperscript{4}\,Iowa State University \\
\textsuperscript{5}\,Woven by Toyota \quad
\textsuperscript{6}\,United Arab Emirates University \\
\textsuperscript{7}\,National Yang Ming Chiao Tung University \quad
\textsuperscript{8}\,University of Macau \\
\textsuperscript{9}\,North Carolina State University \quad
\textsuperscript{10}\,Columbia University \quad
\textsuperscript{11}\,Milestone Systems \\
\textsuperscript{12}\,Xi'an Jiaotong University \quad
\textsuperscript{13}\,Johns Hopkins University
\end{tabular}}

\maketitle

\begin{abstract}
The 10th AI City Challenge, held with ECCV 2026, marks a decade of community benchmarking for intelligent transportation, smart cities, and physical AI. Since its 2017 start with vehicle detection, classification, and tracking, the challenge has grown into a broad benchmark suite for multi-camera perception, multimodal reasoning, synthetic-to-real learning, generative forecasting, and privacy-preserving evaluation. The 2026 edition continued this growth with 325 registered teams, up from 245 in 2025, and participation from 26 countries and regions, up from 15. Its six primary tracks cover multi-camera 3D perception, transportation safety captioning and VQA, traffic anomaly reasoning, text-based person anomaly search, generative traffic video forecasting, and cross-city object detection. Track 3 further includes two out-of-domain leaderboards, submitted as Tracks 7 and 8, for fisheye traffic-violation understanding and pedestrian situated-intent VQA. This paper summarizes the challenge setup, datasets, evaluation protocols, leaderboard results, and workshop papers. Across tracks, successful systems combine foundation models with geometric grounding, retrieval or reranking, synthetic-data design, domain adaptation, and controlled inference.
\end{abstract}

\section{Introduction}
\label{sec:intro}

The AI City Challenge is a recurring benchmark venue for computer vision systems deployed in urban, transportation, and physical AI settings. Its 10th edition, hosted as an ECCV 2026 workshop, marks a decade of growth from the 2017 vehicle detection, classification, and tracking tasks to a wider testbed for multi-camera 3D perception, multimodal reasoning, synthetic-to-real transfer, generative prediction, and privacy-preserving evaluation.

This evolution parallels broader progress in AI. Early tasks reflected deep learning for detection, tracking, and re-identification; recent tasks reflect VQA, foundation models, generative video models, and reasoning systems that must explain events rather than only detect them. The 2026 cross-city object detection track also highlights a persistent limitation of modern AI systems: strong in-domain performance does not guarantee robustness under geographic, camera, and scene-domain shift.

Participation also reached a new high. The 2026 challenge registered 325 teams, compared with 245 in 2025, a growth of about 33\%, and the participant pool represented 26 countries and regions, compared with 15 in the previous year. The breadth of participation is visible across the leaderboards. On the general and public leaderboards, respectively, Track 1 had 29 and 15 teams, Track 2 had 44 and 22, Track 3 had 76 and 27, Track 4 had 47 and 28, Track 5 had 18 and 12, and Track 6 had 39 and 29. The Track 3 out-of-domain leaderboards also drew participation, with 15 general and 8 public teams for FETV fisheye traffic-violation understanding and 15 general and 7 public teams for PSI-VQA pedestrian situated-intent VQA.

This edition therefore introduces a broader set of tasks than a single leaderboard can capture: multi-camera 3D perception, safety-oriented captioning and VQA, anomalous-event reasoning with two OOD leaderboards, text-based person anomaly search, text-conditioned future-frame generation, and cross-city object detection through Milestone Project Hafnia. Together, these tracks form an urban intelligence benchmark suite rather than isolated recognition tasks.

This summary paper follows the structure of recent AI City Challenge overview papers~\cite{Naphade2023AICity7,Wang2024AICity8,Tang2025AICity9}: it first presents the challenge setup, then describes the datasets, evaluation protocols, and leaderboard system, and finally summarizes the track results and the main technical trends in workshop papers. Dataset and benchmark papers associated with the 2026 challenge are discussed in the dataset section, while the results section focuses on teams with identifiable evaluation-system entries.

\section{Challenge Setup}
\label{sec:setup}

The 2026 AI City Challenge was organized around six primary challenge tracks and eight evaluation-server leaderboards, because the main Track 3 task includes two optional out-of-domain evaluations submitted as Tracks 7 and 8 on the evaluation system. Teams registered, requested submission privileges, and submitted predictions to public or general leaderboards~\cite{AICity26Evaluation}. Public leaderboard entries were intended for award-eligible submissions, while the general leaderboards recorded broader participation.

\textbf{Track 1: Multi-Camera 3D Perception.} Participants tracked people, robots, forklifts, pallet trucks, and humanoids across synchronized warehouse cameras. The task required per-frame 3D boxes, class labels, scene-level identities, and world-coordinate localization. Depth was available for training and validation only; hidden real-world testing required RGB-only inference.

\textbf{Track 2: Transportation Safety Understanding and Captioning.} Participants used synthetic Digital Twin WTS data to caption and answer questions about real WTS videos. Each event was divided into behaviorally meaningful phases, and systems generated pedestrian and vehicle captions while also answering multiple-choice questions about position, direction, attention, attributes, and context.

\textbf{Track 3: Anomalous Events in Transportation.} Participants built one unified system that detects, reasons about, and explains anomalous transportation events through binary and multiple-choice questions, open-ended explanation, causal linkage, scene description, temporal description, and summarization. Tracks 7 and 8 extended the same reasoning theme to fisheye traffic violations and pedestrian intent.

\textbf{Track 4: Text-Based Person Anomaly Search.} Participants retrieved real pedestrian images from natural-language descriptions that mention both appearance and behavior. The task is a synthetic-to-real retrieval problem: synthetic image-text pairs provide training signal, while the hidden real-world query-gallery test set measures whether models generalize to realistic abnormal and routine behaviors.

\textbf{Track 5: Generative Traffic Video Forecasting.} Participants generated future frames conditioned on recent history frames and textual descriptions of target future behavior. The task stresses temporal consistency, visual fidelity, and whether a generated traffic sequence remains semantically aligned with safety-critical pedestrian and vehicle behavior.

\textbf{Track 6: Cross-City Object Detection.} Participants trained object detectors through the Hafnia Training-as-a-Service platform, which exposes managed training and benchmarking workflows without allowing direct extraction of the full real-world traffic corpus. The hidden benchmark mixes source- and target-city samples to measure geographic and visual domain shift.

Across the tracks, the challenge combined classical perception metrics such as HOTA and mAP with multimodal language, retrieval, generation, and reasoning metrics. This design encouraged systems that combine foundation models with domain-specific constraints, rather than relying on a single model family across all tasks.

\section{Datasets and Tracks}
\label{sec:datasets}

The 2026 challenge uses six primary tracks and two out-of-domain leaderboards on the shared evaluation system. The primary tracks are described on the challenge web site~\cite{AICity26Track1,AICity26Track2,AICity26Track3,AICity26Track4,AICity26Track5,AICity26Track6}, while Tracks 7 and 8 extend Track 3 to fisheye traffic-violation understanding and pedestrian situated-intent VQA. The following subsections summarize the data settings, hidden-test assumptions, and evaluation focus for each benchmark.

\subsection{Track 1: Physical AI Smart Spaces}

Track 1 extends the PhysicalAI-SmartSpaces benchmark to multi-camera 3D perception in warehouse environments~\cite{AICity26Track1,PhysicalAISmartSpaces}. The training and validation data were generated from a large set of simulated warehouse scenes with synchronized RGB and depth videos, calibrated camera poses, 2D and 3D annotations, top-down maps, and multi-object identities. The hidden test set used real-world video in layouts that differ from the synthetic training scenes, making the task a strict Sim2Real benchmark rather than a closed-world tracking exercise.

\begin{figure}[t]
\centering
\includegraphics[width=0.96\textwidth]{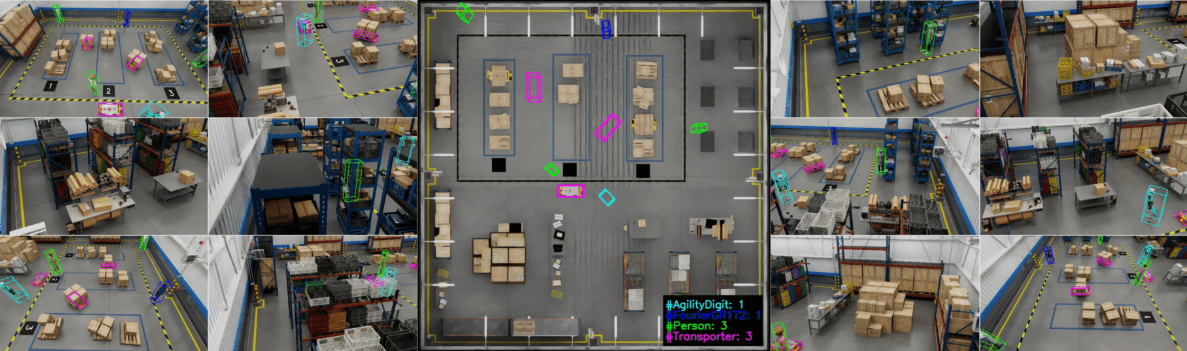}
\caption{Track 1 sample: synchronized multi-camera warehouse views and a top-down layout for RGB-only multi-camera 3D perception.}
\label{fig:track1}
\end{figure}

The central difficulty is the coupling between geometry and association. A system must detect objects in each camera, lift observations into a shared 3D coordinate frame, and maintain identities despite occlusion, similar instances, changing viewpoints, and cluttered warehouse geometry; Fig.~\ref{fig:track1} shows the synchronized views and top-down layout. The RGB-only test restriction prevents direct test-time depth use, so successful approaches rely on learned monocular cues, calibration-aware projection, scene priors, and global association.

\subsection{Track 2: Digital Twin WTS}

\begin{figure}[t]
\centering
\includegraphics[width=0.92\textwidth]{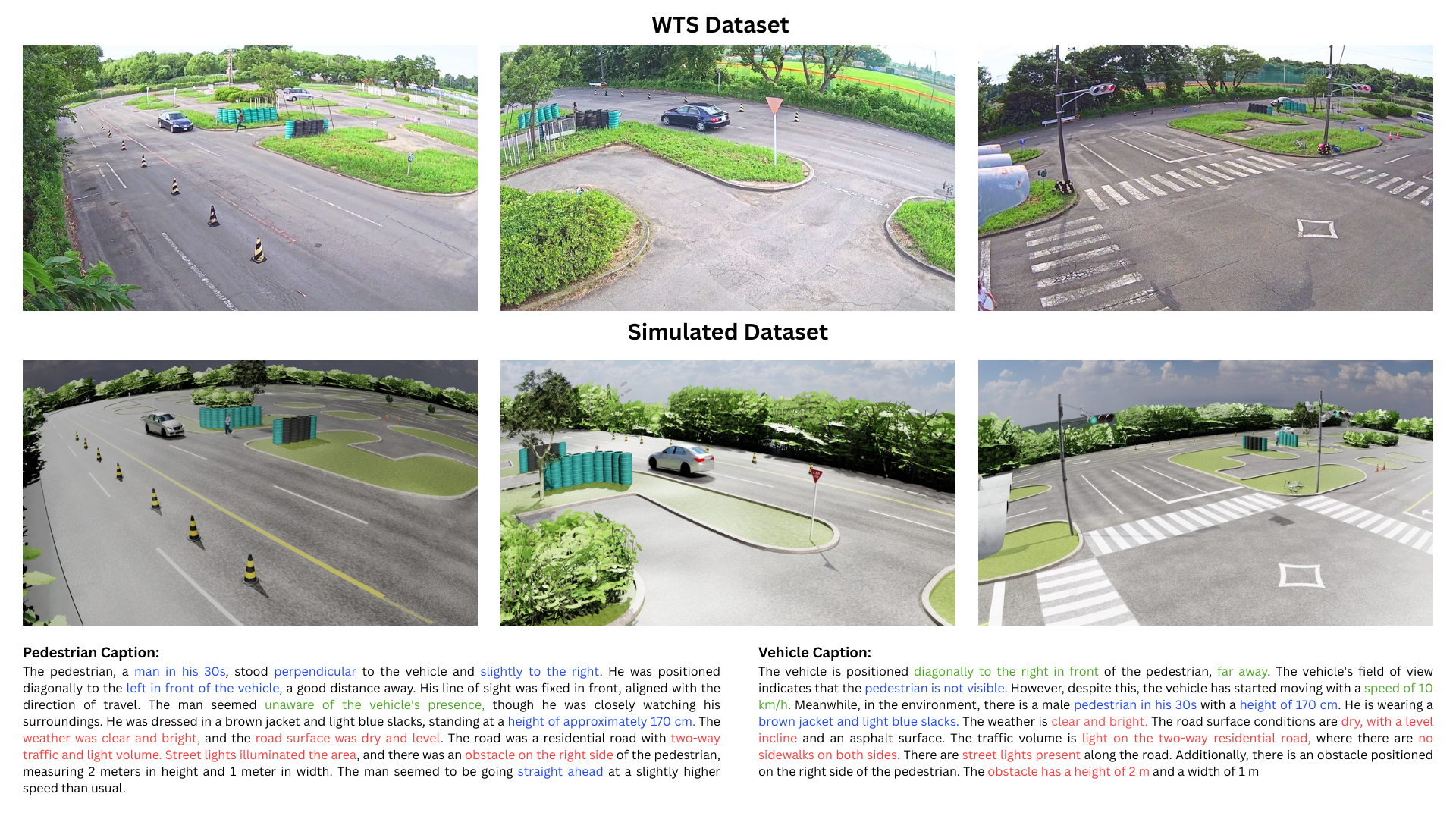}
\caption{Track 2 sample: synthetic Digital Twin WTS scenes paired with real WTS-style views and structured pedestrian/vehicle descriptions.}
\label{fig:track2}
\end{figure}

Track 2 uses the Digital Twin WTS setting for synthetic-to-real traffic safety understanding~\cite{AICity26Track2,SynWTS,AICity26Paper45}. The training and validation data provide synthetic multi-view traffic-safety scenes, annotations, segment-level descriptions, and VQA labels, with representative synthetic and real views shown in Fig.~\ref{fig:track2}. The hidden test set uses real WTS videos, so the benchmark measures whether methods can transfer from controlled synthetic scenes to real camera footage while preserving fine-grained safety semantics.

The task combines two complementary outputs. For captioning, systems describe pedestrian and vehicle behavior across event phases, including movement direction, relative position, gaze, visibility, attributes, weather, road geometry, and traffic context. For VQA, systems answer structured questions that test whether the model has grounded those descriptions in the visual evidence. The track therefore rewards representations that separate stable scene layout from variable appearance and behavior, rather than merely matching surface text patterns.

\subsection{Track 3: Anomalous Events in Transportation}

Track 3 introduces TAR (Traffic Anomaly Reasoning) and TAR-Bench for traffic anomaly reasoning~\cite{AICity26Track3,zhang2026detectionunderstandingtartarbench, PhysicalAITAR}. The TAR training dataset contains 44,040 annotations over 3,670 transportation videos and covers 10 task types, including binary and multiple-choice reasoning, open-ended QA, captioning, causal explanation, scene description, temporal ordering, and summarization. TAR-Bench provides 960 human-curated annotations for 80 held-out clips. Unlike earlier anomaly-detection tasks, this track emphasizes explanation and evidence grounding: a strong model must identify what happened, infer why it happened, and describe the evidence coherently.

\begin{figure}[t]
\centering
\includegraphics[width=0.92\textwidth]{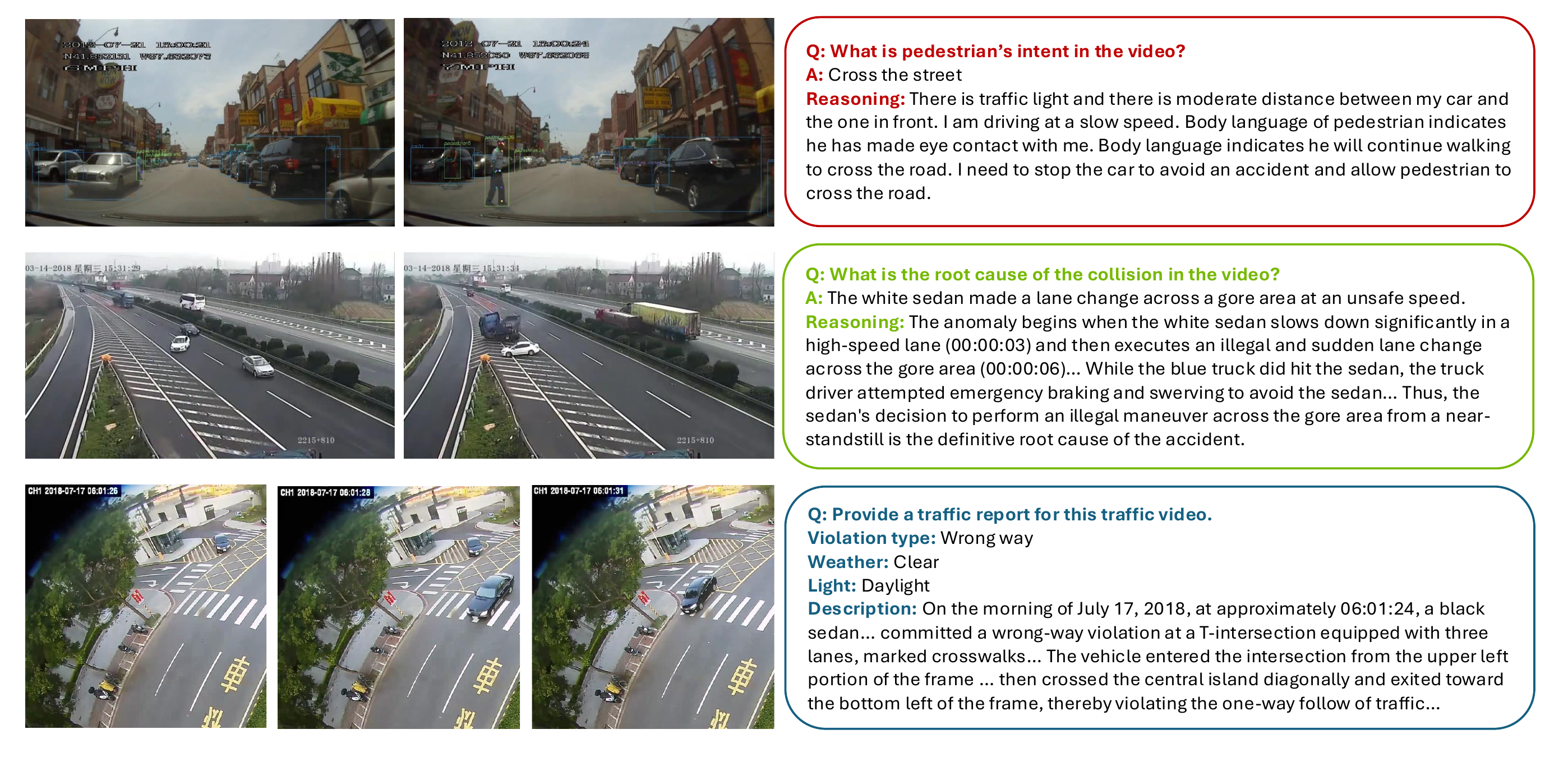}
\caption{Track 3 sample: traffic anomaly reasoning combines video evidence, task-specific questions, event descriptions, and causal explanations.}
\label{fig:track3}
\end{figure}

Two out-of-domain datasets extend Track 3. Track 7 evaluates FETV traffic-violation understanding from fisheye cameras~\cite{AICity26Paper22}, where geometry changes object scale, direction cues, and perspective. Track 8 evaluates PSI-VQA, a pedestrian situated-intent VQA benchmark for ambiguous pedestrian behavior around automated driving~\cite{AICity26Paper35}. Fig.~\ref{fig:track3} illustrates the evidence, question, and explanation structure of the main TAR setting, and the evaluation system reports all three reasoning settings separately as Tracks 3, 7, and 8.

\subsection{Track 4: Pedestrian Anomaly Behavior}

Track 4 uses the Pedestrian Anomaly Behavior (PAB) benchmark for text-based person anomaly search~\cite{AICity26Track4}. The training set is synthetic and provides image-text pairs that describe person appearance, scene context, and action. The real test set contains query descriptions and gallery images. Participants rank gallery images for each query, and the final leaderboard uses retrieval quality on hidden real-world data.

\begin{figure}[t]
\centering
\includegraphics[width=0.92\textwidth]{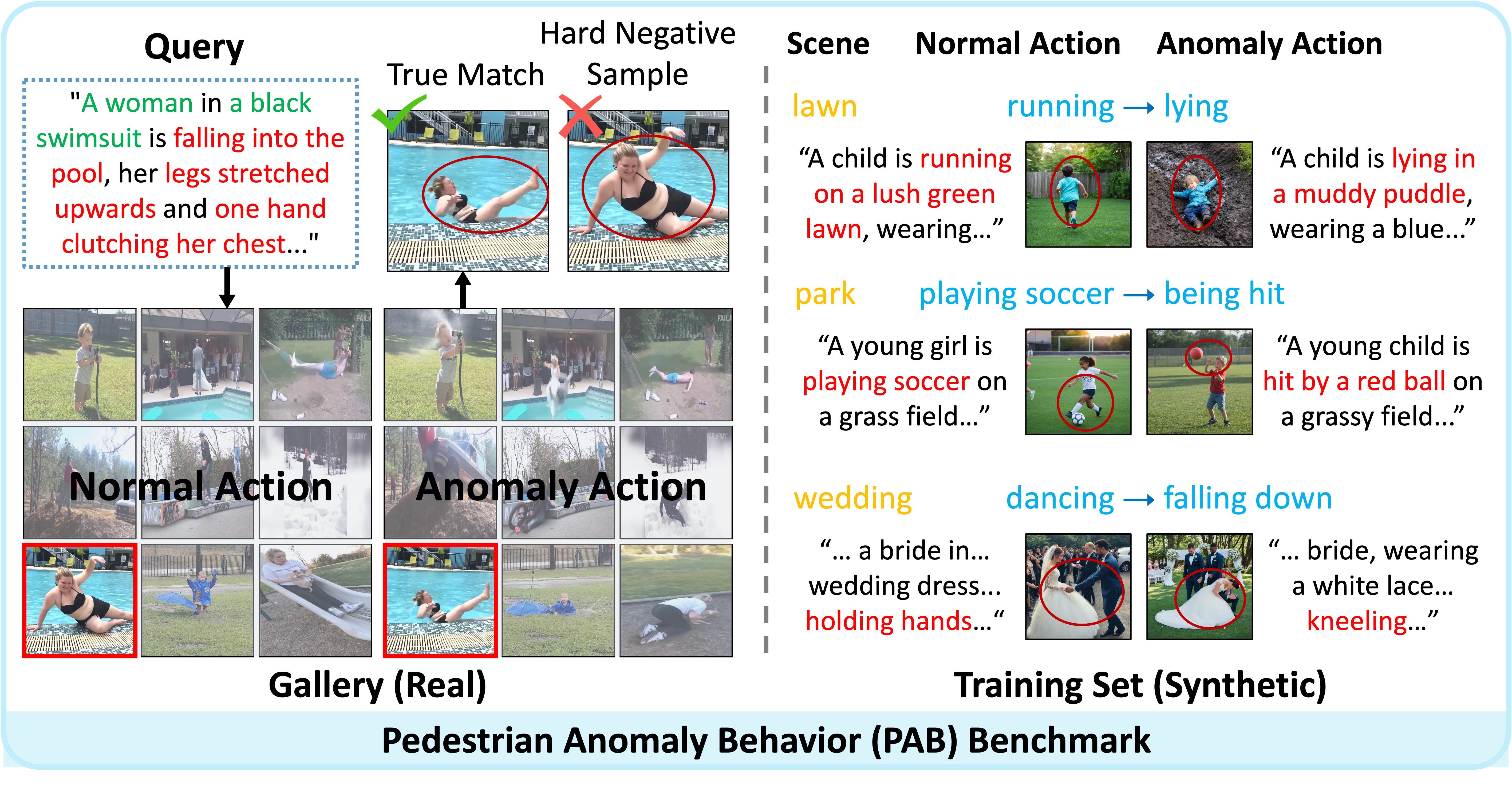}
\caption{Track 4 sample: synthetic PAB training examples and real-gallery retrieval, including hard negatives that share appearance or scene context with the query.}
\label{fig:track4}
\end{figure}

The benchmark is difficult because abnormal behavior is often defined by relationships between action, body pose, scene context, and a natural-language query. Appearance-only retrieval is insufficient: examples such as falling, lying, being hit, or unsafe crossing require action grounding, as the synthetic/real and hard-negative examples in Fig.~\ref{fig:track4} illustrate. Successful methods therefore use text-image contrastive training, action-aware alignment, hard-negative mining, reranking, and query decomposition.

\subsection{Track 5: Traffic Video Forecasting}

\begin{figure}[t]
\centering
\includegraphics[width=0.92\textwidth]{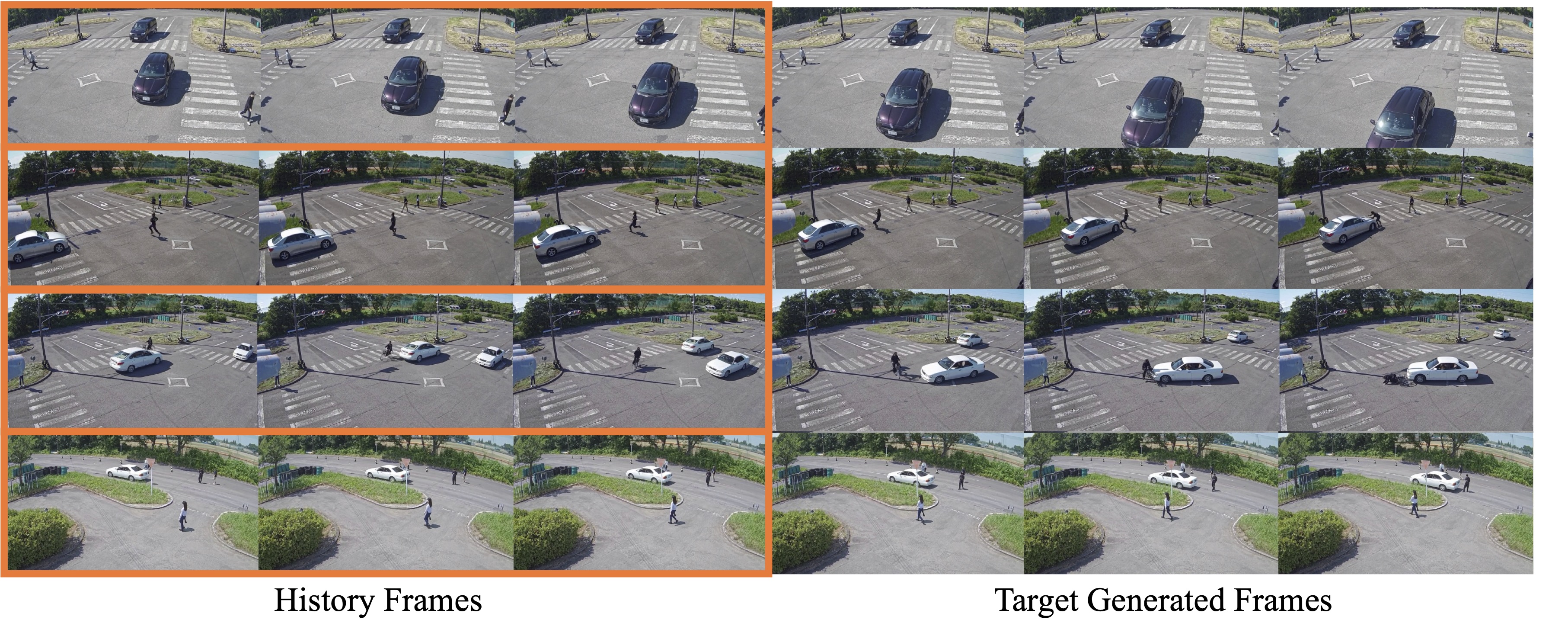}
\caption{Track 5 sample: future-frame generation from history frames and target behavior descriptions in traffic scenes.}
\label{fig:track5}
\end{figure}

Track 5 asks teams to generate future traffic video frames from a short history window and a textual description of expected future behavior~\cite{AICity26Track5}. The task builds on WTS-style traffic scenes but changes the output from analysis to generation, as shown by the history and target-frame example in Fig.~\ref{fig:track5}. Models must synthesize plausible motion, preserve identity and background consistency, and reflect the target pedestrian/vehicle behavior.

This task exposes a different failure mode from captioning and retrieval: a generated video may be sharp but semantically wrong, or text-aligned while losing scene consistency and temporal smoothness. The leaderboard therefore combines low-level fidelity, perceptual, semantic, and video-distribution metrics. Accepted methods typically use diffusion or world-model priors, frame-history conditioning, text-guided planning, and post-generation selection.

\subsection{Track 6: Hafnia Cross-City Detection}

Track 6 uses Hafnia Training-as-a-Service platform~\cite{HafniaTaas} for cross-city object detection~\cite{AICity26Track6,HafniaDataset}. The data consist of real traffic-camera images with object annotations across multiple categories. Unlike tracks where participants directly download the full corpus, this track uses a managed platform for privacy-preserving training, validation, and hidden benchmarking on controlled splits.

\begin{figure}[t]
\centering
\includegraphics[width=0.92\textwidth]{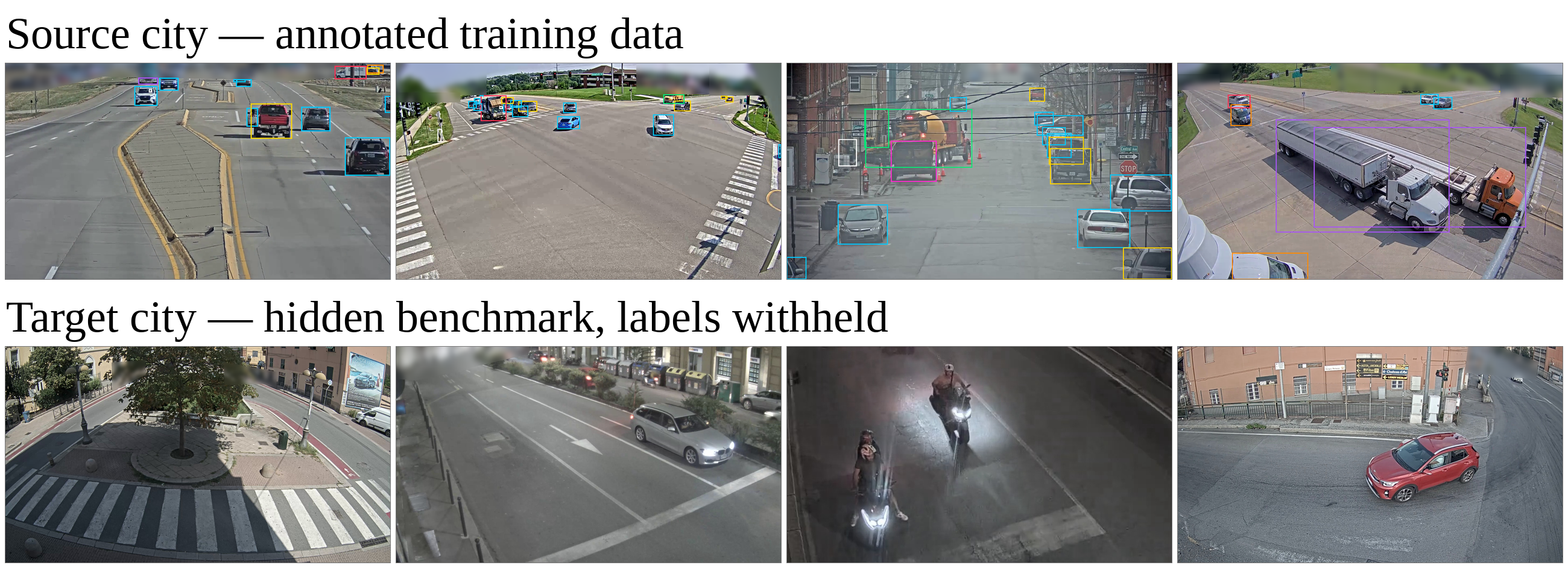}
\caption{Track 6 sample: privacy-preserved traffic-camera images from Milestone Systems (Project Hafnia), with annotated source-city frames on top and hidden target-city frames below.}
\label{fig:track6}
\end{figure}

The track targets a common deployment problem: detectors trained in one city or camera network often lose accuracy when moved to another geography, camera height, lens, weather condition, or traffic pattern, as illustrated in Fig.~\ref{fig:track6}. The hidden benchmark includes source- and target-city samples, rewarding detector design together with class-aware augmentation, resolution management, domain-shift validation, and confidence calibration.

\section{Evaluation Protocols}
\label{sec:evaluation}

The evaluation system exposed public and general leaderboards for each track~\cite{AICity26Evaluation}. During the challenge, scores were computed on a subset of hidden test data and only limited ranking information was shown. After the deadline, final scores were recomputed on full hidden test sets and team names were made public for the workshop summary. Public/general ranks in the results tables use the form ``1/21 (1/58)'', meaning rank 1 among 21 public teams and rank 1 among 58 general teams.

\subsection{Track 1 Evaluation}

Track 1 used 3D HOTA~\cite{HOTA}, which jointly evaluates detection accuracy, association accuracy, and localization quality over 3D tracks. Submissions provided class labels, frame indices, scene-level identities, and 3D boxes or equivalent localization fields in the warehouse coordinate frame. Scores were averaged across classes and scenes. Because the hidden test set was RGB-only, participants could not use test-time depth; online submissions received an additional ranking bonus when they used only current and past frames.

\subsection{Track 2 Evaluation}

Track 2 combined captioning and VQA. Caption quality used BLEU-4~\cite{Papineni2002BLEU}, METEOR~\cite{Banerjee2005METEOR}, ROUGE-L~\cite{Lin2004ROUGE}, and CIDEr~\cite{Vedantam2015CIDEr}; VQA used answer accuracy over structured questions~\cite{Antol2015VQA}. The final score averaged caption and VQA components, making the hidden real WTS test set a synthetic-to-real transfer benchmark for semantic grounding.

\subsection{Track 3 Evaluation}

Track 3 evaluated multi-task traffic anomaly reasoning. The official in-domain TAR-Bench mean was computed from nine scored task types, excluding temporal localization. Closed-form tasks used accuracy-style scoring, while text-generation tasks used BLEU, METEOR, ROUGE, and CIDEr~\cite{Papineni2002BLEU,Banerjee2005METEOR,Lin2004ROUGE,Vedantam2015CIDEr}. The design rewards models that organize evidence over time and generate semantically correct, visually grounded answers.

\subsection{Track 4 Evaluation}

Track 4 used mean Average Precision (mAP), widely used in detection and retrieval benchmarks~\cite{Everingham2010VOC,Lin2014COCO}, for text-to-image retrieval. For each text query, participants returned a ranked gallery list; the score rewards placing correct matches early while suppressing hard negatives. Because training images are synthetic and the test gallery is real, the metric measures both retrieval quality and Sim2Real generalization.

\subsection{Track 5 Evaluation}

Track 5 combined visual, perceptual, and semantic video metrics: PSNR and SSIM~\cite{Wang2004SSIM} for pixel fidelity, LPIPS~\cite{Zhang2018LPIPS} and CLIP-S~\cite{Hessel2021CLIPScore} for perceptual and text-image alignment, and FID~\cite{Heusel2017FID} and FVD~\cite{Unterthiner2018FVD} for distributional realism. The final score rewards realistic future frames that also follow the target behavior described in text.

\subsection{Track 6 Evaluation}

Track 6 used object detection mAP~\cite{Everingham2010VOC,Lin2014COCO} on hidden Hafnia benchmark images. The platform evaluated models on source-city and target-city samples, so the score reflected both in-domain detector quality and robustness to geographic shift. The managed platform also reduced the risk of private-data leakage from the real-world corpus, because participants trained and benchmarked through controlled workflows rather than downloading the full hidden dataset.

\subsection{Tracks 7 and 8 Evaluation}

Tracks 7 and 8 were optional out-of-domain Track 3 leaderboards. Track 7 evaluated FETV traffic-violation understanding from fisheye cameras, while Track 8 evaluated PSI-VQA pedestrian situated-intent questions. Both used VQA- and reasoning-oriented scoring related to the Track 3 protocol~\cite{Antol2015VQA,zhang2026detectionunderstandingtartarbench} and reported final scores separately from the main TAR mean, making transfer robustness visible under fisheye geometry and socially ambiguous pedestrian-intent questions.

Award-candidate teams were required to provide reproducible code and models, and the paper-review process encouraged accepted workshop papers to report the final team names and leaderboard values exactly as shown on the evaluation system. This requirement was especially important for tracks with hidden tests or managed data access, where reproducibility depends on both the submitted model and a clearly documented training and inference pipeline.

\section{Challenge Results}
\label{sec:results}

Tables~\ref{tab:track1}--\ref{tab:track7track8} summarize accepted-paper entries with identifiable evaluation IDs on the 2026 leaderboards. The tables are not intended to replace the full public evaluation system; instead, they connect leaderboard outcomes to accepted workshop papers that describe the corresponding methods. Public ranks and general ranks are both reported because the public leaderboard captures award-eligible entries, while the general leaderboard captures broader participation.

\subsection{Summary for the Track 1 Challenge}

Track 1 attracted methods that combined object detection, camera calibration, 2D-to-3D lifting, and global multi-camera association. The leading accepted systems relied on strong RGB detectors and then used warehouse geometry to reduce cross-view ambiguity. Because the hidden test set did not expose depth, the best submissions treated depth and 3D layout as learned or calibrated priors rather than as direct test-time inputs.

\begin{table}[t]
\centering
\caption{Track 1 leaderboard entries connected to accepted papers.}
\label{tab:track1}
\scriptsize
\begin{tabularx}{\textwidth}{@{}c c l r X@{}}
\toprule
Rank & Team ID & Team & 3D HOTA & Paper \\
\midrule
1/15 (2/30) & 289 & EVA & 56.5447 & \cite{AICity26Paper26}; online=true \\
2/15 (3/30) & 34 & SKKU-AL-T1 & 52.0118 & \cite{AICity26Paper2}; online=true \\
3/15 (4/30) & 130 & Playbox & 38.0105 & \cite{AICity26Paper48}; online=true \\
4/15 (5/30) & 4 & QDTers & 34.1845 & \cite{AICity26Paper49}; online=true \\
5/15 (6/30) & 133 & TU-YMLab & 25.9712 & \cite{AICity26Paper40}; online=true \\
\bottomrule
\end{tabularx}
\end{table}

The accepted Track 1 papers in Table~\ref{tab:track1} show several common design choices. EVA~\cite{AICity26Paper26} led the public leaderboard among accepted papers with geometry-aware tracking. SKKU-AL-T1~\cite{AICity26Paper2} and Playbox~\cite{AICity26Paper48} also performed strongly with online RGB-only pipelines, showing that streaming constraints remain compatible with competitive detection and association. Other papers explored collaborative 2D-3D tracking~\cite{AICity26Paper49} and uncertainty-aware observation construction~\cite{AICity26Paper40}.

\subsection{Summary for the Track 2 Challenge}

Track 2 measured transfer from synthetic Digital Twin WTS data to real WTS videos. The strongest systems used VLM backbones, but the accepted papers also show that off-the-shelf prompting was rarely enough. Teams improved transfer by decomposing the task into scene parsing, phase recognition, caption generation, VQA answering, and answer calibration.

\begin{table}[t]
\centering
\caption{Track 2 leaderboard entries connected to accepted papers.}
\label{tab:track2}
\scriptsize
\begin{tabularx}{\textwidth}{@{}c c l r X@{}}
\toprule
Rank & Team ID & Team & S2 & Paper \\
\midrule
1/22 (1/44) & 47 & Latent Painter - UTE & 60.0853 & \cite{AICity26Paper46}; VL-JEPA \\
2/22 (2/44) & 24 & UIT - Kitchen & 57.3307 & \cite{AICity26Paper30}; VQA \\
3/22 (4/44) & 266 & KZ6 & 56.7949 & \cite{AICity26Paper12}; Qwen-3-VL-8B \\
8/22 (14/44) & 127 & Team KODE & 55.4679 & \cite{AICity26Paper32}; Qwen \\
\bottomrule
\end{tabularx}
\end{table}

The Track 2 results in Table~\ref{tab:track2} suggest that separating scene structure from behavior helps narrow the synthetic-to-real gap. The leading accepted entries used decoupled semantics, V-JEPA style visual features, state-bridging strategies, and modular spatial grounding~\cite{AICity26Paper46,AICity26Paper30,AICity26Paper12,AICity26Paper32}. Methods that represented relative position, pedestrian visibility, vehicle motion, and environmental context were better aligned with the task than generic captioning pipelines.

\subsection{Summary for the Track 3 Challenge}

Track 3 was the largest reasoning leaderboard in the 2026 challenge. It required systems to answer heterogeneous questions about anomalous traffic events and to produce explanations that match visual evidence. The ranking emphasized not only timestamp prediction, binary or multiple choice correctness, but also semantic and causal reasoning soundness.

\begin{table}[t]
\centering
\caption{Track 3 leaderboard entries connected to accepted papers.}
\label{tab:track3}
\scriptsize
\begin{tabularx}{\textwidth}{@{}c c l r X@{}}
\toprule
Rank & Team ID & Team & Mean & Paper \\
\midrule
1/27 (1/76) & 25 & Stellarview AI & 0.6788 & \cite{AICity26Paper19}; Qwen 3.5 \\
- (2/76) & 45 & UOB\&UW Team & 0.6779 & \cite{AICity26Paper39}; Qwen3VL-8B \\
2/27 (5/76) & 60 & FPT AI Vision & 0.6703 & \cite{AICity26Paper53}; Qwen3-VL-8B \\
3/27 (7/76) & 12 & Smart Vision & 0.6669 & \cite{AICity26Paper18}; Qwen \\
10/27 (26/76) & 30 & UWIPL\_ETRI & 0.6185 & \cite{AICity26Paper3}; Qwen \\
15/27 (36/76) & 122 & OptimAI & 0.5880 & \cite{AICity26Paper9}; Qwen \\
16/27 (41/76) & 139 & MR-CAS & 0.5780 & \cite{AICity26Paper52}; GPT \\
24/27 (55/76) & 277 & Korea Drive & 0.4256 & \cite{AICity26Paper4}; Qwen3 \\
\bottomrule
\end{tabularx}
\end{table}

As shown in Table~\ref{tab:track3}, Stellarview AI~\cite{AICity26Paper19} led both the public and general TAR ranking among accepted papers. Other strong submissions used evidence-driven chained reasoning, unified multi-leaderboard agents, metric matching, shared event memory, and structured question routing~\cite{AICity26Paper18,AICity26Paper39,AICity26Paper52,AICity26Paper53,AICity26Paper54}. The accepted papers indicate a shift from simple VLM prompting toward agentic pipelines that first extract visual evidence, then match it to a task-specific answer format.

\subsection{Summary for the Track 4 Challenge}

Track 4 produced high retrieval scores, indicating rapid progress on the PAB synthetic-to-real setting. At the same time, the concentration of strong mAP values made the track sensitive to careful data-use policy and reproducibility checks. The top methods combined strong text-image embeddings with action-specific reranking and hard-negative handling.

\begin{table}[t]
\centering
\caption{Track 4 leaderboard entries connected to accepted papers.}
\label{tab:track4}
\scriptsize
\begin{tabularx}{\textwidth}{@{}c c l r X@{}}
\toprule
Rank & Team ID & Team & mAP & Paper \\
\midrule
1/28 (3/47) & 59 & Xiilab.AIpex & 99.3020 & \cite{AICity26Paper41} \\
3/28 (6/47) & 9 & hiensumi & 98.3535 & \cite{AICity26Paper51} \\
7/28 (9/47) & 27 & VGU AI LAB & 95.4078 & \cite{AICity26Paper10} \\
8/28 (11/47) & 97 & SMART Lab & 94.7815 & \cite{AICity26Paper37} \\
11/28 (14/47) & 76 & HCMUS\_4CentralVN & 93.6715 & \cite{AICity26Paper56} \\
13/28 (15/47) & 93 & SelabHuman & 93.6577 & \cite{AICity26Paper29} \\
15/28 (22/47) & 64 & GenAI4E & 90.9236 & \cite{AICity26Paper55} \\
18/28 (31/47) & 29 & EMBIA & 84.2509 & \cite{AICity26Paper13} \\
\bottomrule
\end{tabularx}
\end{table}

Table~\ref{tab:track4} lists Xiilab.AIpex~\cite{AICity26Paper41} as the top accepted public/general entry in the final decision sheet. Other accepted papers explored late consensus, heterogeneous VLM ensembling, embedding prediction, cross-encoder reranking, global assignment, and action-aligned retrieval~\cite{AICity26Paper10,AICity26Paper13,AICity26Paper29,AICity26Paper37,AICity26Paper51,AICity26Paper55,AICity26Paper56}. The track highlights a broader lesson for text-person search: robust action semantics must be learned together with identity, clothing, and scene context.

\subsection{Summary for the Track 5 Challenge}

Track 5 moved the challenge from recognizing or explaining traffic scenes to generating plausible futures. The leaderboard rewarded systems that maintained scene continuity while following textual descriptions of future behavior. This made the track a natural test of video diffusion models, world models, and text-conditioned forecasting pipelines.

\begin{table}[t]
\centering
\caption{Track 5 leaderboard entries connected to accepted papers.}
\label{tab:track5}
\scriptsize
\begin{tabularx}{\textwidth}{@{}c c l r X@{}}
\toprule
Rank & Team ID & Team & Final & Paper \\
\midrule
1/12 (2/18) & 209 & Qyn & 76.4866 & \cite{AICity26Paper21} \\
2/12 (3/18) & 78 & SSUPER & 76.0385 & \cite{AICity26Paper31} \\
3/12 (4/18) & 47 & Latent Painter - UTE & 75.4302 & \cite{AICity26Paper44} \\
4/12 (6/18) & 83 & CHTTL\_A30 & 74.0544 & \cite{AICity26Paper24} \\
5/12 (7/18) & 39 & VGU\_ai\_lab & 73.3037 & \cite{AICity26Paper15} \\
\bottomrule
\end{tabularx}
\end{table}

The accepted systems summarized in Table~\ref{tab:track5}, including Qyn~\cite{AICity26Paper21}, SSUPER~\cite{AICity26Paper31}, Latent Painter - UTE~\cite{AICity26Paper44}, CHTTL\_A30~\cite{AICity26Paper24}, and VGU\_ai\_lab~\cite{AICity26Paper15}, combined history-frame conditioning with language-guided future descriptions, diffusion or world-model priors, and metric-aware selection. The close scores suggest rapid progress, while the metric suite shows that safety-relevant semantic consistency remains hard to capture with one number.

\subsection{Summary for the Track 6 Challenge}

Track 6 emphasized a deployment problem that is often hidden by conventional object-detection benchmarks: a detector tuned for one camera network can lose accuracy when moved to another city with different viewpoints, object scales, weather, compression artifacts, and traffic composition. The Hafnia platform made this setting more realistic by using managed training and evaluation workflows, so teams had to improve domain robustness without directly extracting the full real-world corpus. This makes the track closer to practical cross-site deployment, where data governance and distribution shift are handled together.

The accepted papers in Table~\ref{tab:track6} show several complementary approaches to cross-city detection. SKKU-AL-T1~\cite{AICity26Paper20} led the accepted-paper entries with a strategy focused on pre-training and augmentation for zero-shot transfer. BIT-ODL~\cite{AICity26Paper23} used evidence-conditioned multi-source pretraining, while BK2\allowbreak TheFuture~\cite{AICity26Paper28} emphasized targeted augmentation and class-aware inference. The lyx submission~\cite{AICity26Paper8} explored two-stage fusion and scale-aware geometric refinement. Although the final mAP values are lower than the retrieval scores in Track 4, this should be interpreted in light of the task: Track 6 evaluates object localization, class recognition, and confidence ranking under cross-city shift on real imagery. Across the accepted methods, the recurring theme is that detector architecture alone is not sufficient. Successful systems also manage resolution, category imbalance, source-domain bias, validation under distribution shift, and confidence calibration for target-city scenes.

\begin{table}[t]
\centering
\caption{Track 6 leaderboard entries connected to accepted papers.}
\label{tab:track6}
\scriptsize
\begin{tabularx}{\textwidth}{@{}c c l r X@{}}
\toprule
Rank & Team ID & Team & mAP & Paper \\
\midrule
1/25 (1/35) & 34 & SKKU-AL-T1 & 0.4753 & \cite{AICity26Paper20}\\
2/25 (2/35) & 265 & BIT-ODL & 0.4281 & \cite{AICity26Paper23}\\
4/25 (5/35) & 261 & BK2TheFuture & 0.4169 & \cite{AICity26Paper28}\\
- (7/35) & 315 & lyx & 0.4060 & \cite{AICity26Paper8}\\
\bottomrule
\end{tabularx}
\end{table}

\subsection{Summary for the Track 7 and Track 8 OOD Challenges}

Tracks 7 and 8 extended the Track 3 reasoning task into two out-of-domain settings. Track 7 used FETV traffic-violation understanding from fisheye cameras, where wide-angle projection changes object shape, direction cues, lane geometry, and motion interpretation. Track 8 used PSI-VQA for pedestrian situated-intent reasoning, where answers depend on subtle social signals, occlusion, ambiguity, and the viewpoint of an automated-driving agent. These tracks are therefore not merely auxiliary leaderboards; they test whether traffic reasoning systems transfer from in-domain CCTV anomaly clips to different sensing geometries and interaction questions.

Table~\ref{tab:track7track8} summarizes accepted-paper entries on both OOD leaderboards. UniTraffic~\cite{AICity26Paper3} ranked first among accepted public entries on both settings, suggesting that evidence-centric agentic reasoning can transfer across anomaly, violation, and intent tasks. UniTraffic-Agent~\cite{AICity26Paper52} and Korea Drive~\cite{AICity26Paper4} also performed strongly with unified or task-routed video-language reasoning. TAU-Agent~\cite{AICity26Paper39} appears in the general rankings and reflects another design pattern: retrieval-augmented reasoning can help reuse event evidence across related but shifted traffic-understanding tasks. The Track 7 scores are close among the top accepted entries, indicating that fisheye violation understanding remains sensitive to spatial evidence extraction. Track 8 has a wider score range, consistent with the added difficulty of socially ambiguous pedestrian-intent questions. Together, these OOD leaderboards complement TAR by measuring robustness, not only in-domain reasoning accuracy.

\begin{table}[t]
\centering
\caption{Track 7 and Track 8 OOD leaderboard entries connected to accepted papers.}
\label{tab:track7track8}
\scriptsize
\begin{tabularx}{\textwidth}{@{}c c c l r X@{}}
\toprule
Track & Rank & Team ID & Team & Final & Paper \\
\midrule
\multicolumn{6}{@{}l}{\textit{Track 7: FETV traffic-violation understanding}} \\
\midrule
7 & 1/8 (1/15) & 30 & UWIPL\_ETRI & 0.4891 & \cite{AICity26Paper3} \\
7 & 2/8 (3/15) & 139 & MR-CAS & 0.4884 & \cite{AICity26Paper52} \\
7 & 3/8 (5/15) & 277 & Korea Drive & 0.4634 & \cite{AICity26Paper4} \\
7 & - (12/15) & 45 & UOB\&UW Team & 0.3998 & \cite{AICity26Paper39} \\
\midrule
\multicolumn{6}{@{}l}{\textit{Track 8: PSI-VQA pedestrian situated-intent reasoning}} \\
\midrule
8 & 1/7 (2/15) & 30 & UWIPL\_ETRI & 70.6397 & \cite{AICity26Paper3} \\
8 & - (5/15) & 73 & University of Washington & 67.9275 & \cite{AICity26Paper39} \\
8 & 4/7 (8/15) & 139 & MR-CAS & 64.4161 & \cite{AICity26Paper52} \\
8 & 5/7 (9/15) & 277 & Korea Drive & 57.0400 & \cite{AICity26Paper4} \\
\bottomrule
\end{tabularx}
\end{table}

\section{Discussion and Conclusion}
\label{sec:conclusion}

Across a decade, the AI City Challenge has grown from vehicle detection, classification, tracking, and counting to broader urban intelligence. In 2026, 325 teams from 26 countries and regions evaluated systems across six primary tracks and two OOD leaderboards: multi-camera 3D perception, synthetic-to-real safety understanding, anomaly reasoning, text-based search, generative forecasting, and cross-city detection. The track portfolio reflects how the field has moved from isolated CV predictions toward systems that combine geometry, language, temporal evidence, generation, and domain adaptation.

The 2026 results also show that progress is uneven in productive ways. Some teams achieved strong retrieval and reasoning scores with foundation-model pipelines, while the cross-city detection, RGB-only 3D perception, video forecasting, and OOD reasoning settings exposed remaining gaps in transfer, calibration, and reproducibility. This is a useful outcome for a benchmark: the best submissions identify practical solution patterns, while the remaining failures define research directions for future editions. We expect subsequent challenges to emphasize reproducible end-to-end systems, more explicit OOD testing, privacy-preserving evaluation platforms, and interaction with related programs such as IARPA Video LINCS~\cite{VideoLINCS}.

Across the accepted papers, three patterns are especially visible: many high-ranking entries used modular pipelines, the strongest Sim2Real and cross-city methods treated domain shift as a first-class constraint, and the multimodal tracks showed that language is useful only when grounded in visual evidence. Together, these patterns connect the 2026 tracks into a shared research agenda for deployable urban AI.

\section*{Acknowledgments}

Rama Chellappa was supported by the VideoLINCS program. This research is based upon work supported in part by the Office of the Director of National Intelligence (ODNI), Intelligence Advanced Research Projects Activity (IARPA), via 56000026C0026. The views and conclusions contained herein are those of the authors and should not be interpreted as necessarily representing the official policies, either expressed or implied, of ODNI, IARPA, or the U.S. Government. The U.S. Government is authorized to reproduce and distribute reprints for governmental purposes notwithstanding any copyright annotation therein.

\bibliographystyle{splncs04}
\bibliography{main,aicity26}

\end{document}